\documentclass{article}

\usepackage[eandd, nonanonymous, preprint]{neurips_2026}

\usepackage[utf8]{inputenc}
\usepackage[T1]{fontenc}
\usepackage{hyperref}
\usepackage{url}
\usepackage{booktabs}
\usepackage{amsfonts}
\usepackage{nicefrac}
\usepackage{microtype}
\usepackage{xcolor}
\usepackage{adjustbox}
\usepackage{multirow}
\usepackage{makecell}
\usepackage{graphicx}
\usepackage{float}
\graphicspath{{image/}}
\usepackage[table]{xcolor}
\usepackage{pifont}
\usepackage{enumitem}
\newcommand{\cmark}{\textcolor{green!60!black}{\ding{51}}}
\newcommand{\xmark}{\textcolor{red}{\ding{55}}}
\newcommand{\pmark}{\textcolor{orange}{\ding{115}}}
\usepackage{amsmath}

\title{AgentPerfBench: A Benchmarking and Evaluation Suite for Inference Performance of Agentic LLMs}

\author{%
  \normalfont
  Cheuk Hang Lau\textsuperscript{1} \quad
  Zeyu Cao\textsuperscript{2} \quad
  Kevin Wong Cheuk Yin\textsuperscript{1} \quad
  Yao Lai\textsuperscript{2} \\[0.15em]
  Haoran Wu\textsuperscript{2} \quad
  Nicholas D. Lane\textsuperscript{2} \quad
  Robert D. Mullins\textsuperscript{2} \quad
  Ilia Shumailov\textsuperscript{3} \quad
  Yiren Zhao\textsuperscript{1} \\[0.45em]
  \textsuperscript{1}Imperial College London \quad
  \textsuperscript{2}University of Cambridge \quad
  \textsuperscript{3}University of Oxford
}

\begin{document}

\maketitle

\begin{abstract}
The optimization of LLM serving engines, such as vLLM and SGLang, as well as the development of new AI hardware, is largely benchmark-driven: optimizations, scheduling policies, hardware and system designs are all selected based on representative workloads. However, a significant mismatch has emerged in the agentic era. Existing benchmarks primarily focus on simple single-turn chatbot workloads. In practice, LLM applications are increasingly agentic: coding agents, automated terminal execution systems, and tool-use agents issue multi-turn requests with growing context lengths across sessions that can span hundreds of interaction steps. To this end, we introduce \textsc{AgentPerfBench}, a new benchmark suite for agentic inference. It uses real traces from widely used agentic benchmarks, such as SWE-Bench and TerminalBench, alongside standard chat baselines. This enables benchmarking of models on multi-turn tasks involving tool calling, skill utilization, and increasing context lengths. AgentPerfBench also samples from empirical distributions of input length, output length, and turn count derived from the real traces, generating representative synthetic profiles for cheap and accurate measurements on new hardware. In addition, we further find that several existing benchmarks fail to accurately reflect real hardware performance for two key reasons: 1) they do not account for realistic context-length growth, and 2) they measure inference performance without operating at hardware saturation. We discuss these issues in detail and provide rich kernel-level Nsight Compute (NCU) traces to construct a new multi-dimensional roofline model that captures hardware-system limitations in both memory bandwidth and memory capacity footprint. The benchmarking suite then includes automated scripts to identify potential bottleneck conditions on emerging hardware when evaluated with diverse agentic traces. Together, these contributions quantify the chat-to-agentic gap in current inference benchmarks and characterise per-kernel GPU resource utilisation via roofline analysis. 
\end{abstract}

\section{Introduction}
\label{sec:intro}

Agentic inference has driven a dramatic increase in the volume of LLM inference workloads. This makes LLM inference a primary workload for future datacenters and emerging hardware platforms. Agentic workloads, on the other hand, differ significantly from those captured by current inference benchmarks~\citep{mlperf_v51,vllm,sglang,sharegpt}, which primarily focus on simple single-turn, chatbot-style interactions.
These benchmarks mainly have three gaps towards real-world agentic inference workload. Firstly, no public benchmark measures in the agentic, multi-turn, tool-using regime. Coding and computer-use agents issue long-context multi-turn requests that neither MLPerf, vLLM, nor SGLang exercises in their native benchmark. Secondly, existing benchmarks set ISL (input sequence length) and OSL (output sequence length) as fixed user-supplied variables rather than deriving them from real workload traces. For example, InferenceX~\citep{inferencex} fills requests with random tokens at arbitrary fixed lengths, discarding the statistical structure of real workloads. Last but not least, every existing public serving-benchmark harness reports a single operating point: MLPerf Server~\citep{mlperf} fixes open-loop Poisson arrivals at a pre-chosen rate; vLLM and SGLang default to a fixed in-flight-request level, and none sweeps the full load curve to locate saturation to reflect the real capability of the underlying hardware system.

In contrast, AgentPerfBench is a benchmark suite and analysis framework designed to address these gaps. AgentPerfBench organizes workloads into agentic profiles, where each profile captures key workload characteristics: per-turn ISL, per-turn OSL, and turn-count distributions measured from real agentic traces (e.g. coding agents, terminal and computer-use agents). We then instantiate requests by sampling from these profiles and filling them with random tokens at the sampled lengths to stress test the underlying hardware to its maximum capability.\footnote{Sequence length and batch scheduling determine TTFT and TPOT; token content does not. Random tokens at distribution-matched lengths produce the same serving behavior as real tokens at the same lengths.} AgentPerfBench also profiles individual Compute Unified Device Architecture (CUDA) kernels using Nsight Compute (NCU) under these different agentic profiles and maps each workload onto a multi-dimensional roofline model to identify hardware bottlenecks. We follow the formalization of \citet{zhao2026heterogeneous} to quantify both memory-bandwidth and memory-capacity limitations, thus providing the practical performance constraints of different hardware platforms when serving different agentic workloads.

AgentPerfBench can provide significant benchmarking results compared to canonical benchmarking suites (e.g., MLPerf Inference~\citep{mlperf_v51}). For example, for LLaMA-3.1-70B on H100, switching from chat to coding-agent workloads increases time-to-first-token (TTFT) by 4.8$\times$ (239\,ms to 1139\,ms) and time-per-output-token (TPOT) by 1.5$\times$ (19\,ms to 28\,ms) at saturation. Multi-turn sessions compound the TTFT gap: SWE-Bench agents over 88 turns push context past 32K tokens (the model's context limit), driving per-turn median TTFT to 433\,ms versus 62\,ms for chat, a 7.0$\times$ increase; per-turn TPOT remains stable at 16--18\,ms across both workloads (Sec.~\ref{sec:results}).
\textsc{AgentPerfBench} makes three contributions:

\begin{enumerate}
    \item \textbf{Agentic workload driven benchmarking.} We provide a benchmark that captures diverse agentic workloads, including coding, terminal, and computer-use agents. These workloads involve multi-turn conversations, long-context tasks, and heavy tool use, providing a more representative measurement of today’s agentic LLM inference landscape.
    \item \textbf{Saturation-based measurements.} We examine the measurement techniques used in popular benchmarks and propose saturation-based evaluation: measuring performance under the maximum sustainable number of concurrent user requests and reporting results only after hardware performance plateaus. 
    We quantify how inaccurate measurement methodologies can lead to measurements under-reporting true hardware capacity and show that saturation-based measurements better reflect the true maximum capability of the underlying hardware.

    \item \textbf{Kernel-level multi-dimensional roofline analysis.}
    We release per-kernel Nsight Compute (NCU) profiles across diverse agentic workloads and provide a multi-dimensional roofline analysis on representative kernels — covering both memory bandwidth and memory-capacity footprint — to quantify the limiting factors of performance.
\end{enumerate}

We release 3{,}000+ benchmark results, 140{,}000+ per-kernel Nsight Compute (NCU) profiling records across four GPU platforms and 11 model architectures (Appendix Tables~\ref{tab:hardware} and~\ref{tab:models}), and the benchmarking/profiling tooling as an open-source codebase, all under an open-source license.\footnote{Dataset and code: \url{[anonymized for review]}.} Section~\ref{sec:background} reviews LLM inference, the roofline model, and existing serving benchmarks. Section~\ref{sec:method} defines the benchmarking methodology, workload surface, and NCU profiling pipeline. Section~\ref{sec:results} presents the benchmark sweep, serving hardware roofline, and the kernel-level regime shift.

\section{Background}
\label{sec:background}

\paragraph{Inference benchmarks.}
Table~\ref{tab:comparison} compares existing systems across nine capabilities; no prior system provides agentic workloads, multi-turn sessions, NCU profiling, open-source traces, or saturation-based measures.
MLPerf Inference~\citep{mlperf} provides standardized throughput and latency measurements across four scenarios (single-stream, multistream, server, offline) with 99\% confidence intervals over 270K queries. The v5.1 LLM suite~\citep{mlperf_v51} extends coverage to summarization and long-context reasoning from 1{,}024-token Q\&A (LLaMA~2~70B on OpenOrca) up to 128K-token document tasks (LLaMA~3.1~405B), but remains single-turn with no multi-step tool use.
The most commonly used recipes that are widely used in the community are the benchmark harnesses bundled in vLLM~\citep{vllm} and SGLang~\citep{sglang} target ShareGPT~\citep{sharegpt} or fixed-length random-token traces at a single operating point. While both expose request-rate and max-concurrency knobs, neither's standard recipe sweeps to saturation by design, replays multi-turn sessions, or uses agentic traces. InferenceX~\citep{inferencex} provides a public dashboard of hardware comparisons but uses fixed-length synthetic tokens that do not match agentic trace distributions. ML.ENERGY~\citep{mlenergy} benchmarks energy consumption across 40 models and 6 tasks, achieving up to 44\% energy savings via Pareto optimization, but does not measure TTFT or provide kernel-level analysis.

\paragraph{Prefill-decode separation and the roofline model.}
Autoregressive LLM inference proceeds in two phases: \textit{prefill}, which computes on all $L$ input tokens simultaneously within each layer, and \textit{decode}, which generates tokens one at a time, each loading the accumulated Key-Value (KV) cache. \citet{zhao2026heterogeneous} formalize the separation with two quantities. For a single linear layer $Y{=}WX$ ($W \in \mathbb{R}^{m \times d}$, $X \in \mathbb{R}^{d \times L}$), \textit{operational intensity (OI)} metric $\text{OI} = 2mdL\,/\,[b_p\,(md + dL + mL)]$ gives the FLOPs-per-byte ratio that separates compute-bound from memory-bound execution, where $b_p = 2$ bytes per element in bfloat16 (bf16) precision. As $L$ grows, OI approaches the compute-bound asymptote $2md/[b_p\,(m+d)]$; prefill ($L = \text{ISL}$) pushes toward this ceiling while each decode step ($L = 1$) remains near the memory-bound floor. They also define a \textit{capacity footprint (CF)} metric $\text{CF} = b_p\,[2dL + md/B]$ that gives the DRAM bytes per concurrent request, where $B$ is the batch size sharing the weight matrix (weight share plus KV cache).

\paragraph{Workload-dependent OI and the snowballing effect.}
A coding agent with ISL ${\approx}\,$17K pushes prefill OI toward the compute-bound asymptote $2md/(m{+}d)$, while chat with ISL in the hundreds remains near the memory-bound floor~\citep{zhao2026heterogeneous}. In multi-turn sessions, context accumulates across turns: each turn appends the prior response and the next user message, growing ISL and therefore prefill OI with every step. \citet{zhao2026heterogeneous} term this the \textit{snowballing effect}; by the final turn of a long coding-agent session, accumulated context reaches up to 120{,}000 tokens. When CF exceeds the DRAM capacity of a single accelerator, serving requires either tensor parallelism or KV-cache paging.

\begin{table}[t]
\caption{\textbf{Comparison of representative LLM inference simulators and benchmarks.} \cmark\ denotes full support, \xmark\ no support, and \pmark\ partial support. MoE denotes mixture-of-experts model support. InferenceX (a leaderboard dashboard), Sarathi-Serve, and Splitwise (scheduling contributions without standalone prediction accuracy) are discussed in text but omitted; Appendix Figure~\ref{fig:methodology_comparison} provides the full feature-coverage grid.}
\label{tab:comparison}
\centering
\small
\setlength{\tabcolsep}{4pt}
\renewcommand{\arraystretch}{1.1}
\begin{adjustbox}{max width=\textwidth}
\begin{tabular}{l ccccccc}
\toprule
\textbf{Feature} & \textbf{LLMCompass} & \textbf{Vidur} & \textbf{LLMServSim\,2} & \textbf{DistServe} & \textbf{MLPerf} & \textbf{ML.ENERGY} & \textbf{Ours} \\
 & \scriptsize\citep{llmcompass} & \scriptsize\citep{vidur} & \scriptsize\citep{llmservingsim} & \scriptsize\citep{distserve} & \scriptsize\citep{mlperf} & \scriptsize\citep{mlenergy} & \\
\midrule
\rowcolor{gray!15}
\multicolumn{8}{l}{\emph{Measurement}} \\
\quad TTFT/TPOT/ITL & \cmark & \cmark & \cmark & \cmark & \xmark & \xmark & \cmark \\
\quad TP                & \cmark & \cmark & \cmark & \pmark & \xmark & \xmark & \cmark \\
\quad NCU profiling                    & \xmark & \xmark & \xmark & \xmark & \xmark & \xmark & \textbf{\cmark} \\
\quad Agentic workloads               & \xmark & \xmark & \xmark & \xmark & \xmark & \xmark & \textbf{\cmark} \\
\quad Multi-turn                              & \xmark & \xmark & \xmark & \xmark & \xmark & \xmark & \textbf{\cmark} \\
\midrule
\rowcolor{gray!15}
\multicolumn{8}{l}{\emph{Model architecture}} \\
\quad Dense                                   & \cmark & \cmark & \cmark & \cmark & \cmark & \cmark & \cmark \\
\quad MoE                                     & \pmark & \xmark & \cmark & \xmark & \xmark & \cmark & \cmark \\
\midrule
\rowcolor{gray!15}
\multicolumn{8}{l}{\emph{Analysis}} \\
\quad Open-source traces                & \xmark & \xmark & \xmark & \xmark & \xmark & \xmark & \textbf{\cmark} \\
\quad Saturation-based measures                       & \xmark & \xmark & \xmark & \xmark & \xmark & \xmark & \textbf{\cmark} \\
\bottomrule
\multicolumn{8}{@{}l}{\scriptsize $^{\dagger}$Single-layer prefill composition error; other rows report end-to-end serving error.} \\
\end{tabular}
\end{adjustbox}
\vspace{-15pt}
\end{table}

\section{Methodology}
\label{sec:method}

Section~\ref{sec:method} describes the AgentPerfBench structure. We construct workload profiles from real agentic benchmarks traces (covering chat, coding, terminal, and computer-use agents) in two forms: exact trace replay and synthetic profiles fitted to the same per-turn statistics. We then run a closed-loop serving harness across dense and Mixture-of-Experts (MoE) models on the GPU platforms (summarized in Appendix Table~\ref{tab:hardware}), recording performance metrics. Finally, we collect NCU traces of the same workloads to ground end-to-end serving behaviour in per-kernel compute, memory bandwidth, and capacity footprint limits.

\subsection{Workload Surface}
\label{sec:taxonomy}

Table~\ref{tab:profiles} defines six canonical workload profiles spanning chat, coding-agent, terminal-agent, and computer-use sessions with empirical bases from ShareGPT-style chat~\citep{sharegpt}, SWE-Bench~\citep{swebench}, TerminalBench~\citep{terminalbench}, and OSWorld~\citep{osworld}. These canonical names are used throughout the results. 
The released dataset instantiates these profiles in two complementary forms.
Trace replay profiles preserve the exact recorded requests from the source traces. 
Distributional synthetic profiles sample multi-turn requests from per-turn ISL, OSL, and turn-count distributions fitted to those traces, producing representative workloads that can be reissued across models, hardware, backends, and concurrency settings, without the runtime cost of full replay.
Appendix Table~\ref{tab:released_profile_ids} maps the canonical workload names to their released profile IDs.

\begin{table}[t]
\caption{Canonical workload profiles used to define the workload surface. ISL and OSL are profile-level median tokens per request. Session turns is the median number of turns per session; single-turn rows comprise repeated independent requests, while multi-turn rows advance turn-by-turn with accumulating context. Profile IDs used in dataset are listed in Appendix Table~\ref{tab:released_profile_ids}.}
\label{tab:profiles}
\centering
\small
\begin{adjustbox}{max width=\linewidth}
\begin{tabular}{@{}lllrrr@{}}
\toprule
\textbf{Workload} & \textbf{Empirical basis} & \textbf{Session type} & \textbf{ISL} & \textbf{OSL} & \textbf{Session turns} \\
\midrule
Single-turn chat & ShareGPT & single-turn & 187 & 299 & 1 \\
Chatbot session & ShareGPT & multi-turn & 1{,}291 & 169 & 10 \\
Coding agent & SWE-Bench & multi-turn & 9{,}995 & 32 & 85 \\
Terminal agent & TerminalBench & multi-turn & 7{,}811 & 31 & 61 \\
Computer-use agent & OSWorld  & multi-turn & 1{,}399 & 85 & 8 \\
\bottomrule
\end{tabular}
\end{adjustbox}
\end{table}

\subsection{Synthetic Trace Validation}
\label{sec:trace_fidelity}

We validate that synthetic distributional profiles reproduce the serving behavior of trace replay before using them for hardware conclusions. With source sessions, model, GPU, backend, concurrency, context cap, and vLLM prefix-cache settings held fixed, we compare median relative error for TTFT, TPOT, and E2EL over both aggregate requests and fixed turn-depth bins.

Across the agentic profiles, token-count matching captures the general serving behavior, but the coding-agent validation shows that it is not sufficient: the generator must also match code-like token structure and the serving-visible prefix structure detected by vLLM automatic prefix caching (APC). Synthetic distributional profiles in the paper use the resulting prefix-aware configuration; Appendix~\ref{sec:appendix_dist_replay_validation} reports the coding-agent ablation.

\subsection{Benchmarking Framework}
\label{sec:framework}
All benchmarks target OpenAI-compatible endpoints and consume streaming responses. For each request, we record time-to-first-token (TTFT), time-per-output-token (TPOT), inter-token latency (ITL), end-to-end latency (E2EL), input and output sequence length (ISL/OSL), and client-side scheduling metadata. Multi-turn runs also record session and turn identifiers, previous and total context length, new-prefill and cached-context tokens, cache-hit rate, block-aligned cache estimates, uncached prefix tails, and per-turn latency/token/cache summaries.

We evaluate serving with a \textit{closed-loop concurrency} sweep. At concurrency $C$, the benchmark harness admits at most $C$ requests to the endpoint at a time. When one request completes, the next request or session turn is dispatched. For multi-turn profiles, each slot advances a session in turn order, preserving the session-context snowballing that makes agentic prompts grow across turns, increasing prefix-cache pressure.

For request $i$, $\mathrm{TTFT}_i$ is the latency from dispatch to the first streamed token. $\mathrm{ITL}$ records the subsequent inter-token gaps, and $\mathrm{TPOT}_i$ is their mean, excluding the first generated token. End-to-end latency is the time from dispatch to the final token, with the request-level identity
\[
\mathrm{E2EL}_i
=
\mathrm{TTFT}_i
+
\mathrm{TPOT}_i \cdot \max(OSL_i - 1, 0),
\]
where $OSL_i$ is the number of generated output tokens. We report mean, median, p90, and p99 summaries for latency metrics, plus request throughput and input/output-token throughput. \textit{Request throughput} (req/s) is the headline capacity metric; we prefer it to output-token throughput (tok/s), which hides the agentic gap because high-ISL requests demand long prefills but produce few output tokens. \textit{Saturation} is the concurrency $C^{\star}$ at which req/s ceases to rise ($|d(\text{req/s})/dC| < \epsilon$); we sweep $C$ to locate it for every workload, with the TTFT knee and KV-cache fill fraction corroborating $C^{\star}$ but not replacing it.

\subsection{Kernel-level Profiling}
\label{sec:ncu_method}

We decompose each decoder layer into constituent operations and compute component-level FLOPs and DRAM bytes from model dimensions, precision, tensor-parallel degree, active token count, and batch size. For component $j$, the operational intensity is $\mathrm{OI}_j=\mathrm{FLOPs}_j/\mathrm{DRAMBytes}_j$, and the roofline throughput ceiling is:
\[
  P^{\mathrm{roof}}_j = \min\left(P_{\mathrm{peak}},\; \mathrm{OI}_j \times \mathrm{BW}_{\mathrm{peak}}\right),
\]
where $P_{\mathrm{peak}}$ is peak compute throughput (FLOP/s) and $\mathrm{BW}_{\mathrm{peak}}$ is peak memory bandwidth (byte/s) for the target GPU. We convert the ceiling into a component latency lower bound, $t^{\mathrm{roof}}_j=\mathrm{FLOPs}_j/P^{\mathrm{roof}}_j$, and aggregate component costs to obtain a layer-level roofline bound. The ridge point $\mathrm{OI}_{\mathrm{ridge}}=P_{\mathrm{peak}}/\mathrm{BW}_{\mathrm{peak}}$ is the reference used in Section~\ref{sec:kernel} to interpret whether a component is limited by the compute ceiling or the memory-bandwidth slope.

We ground this analytical decomposition with two NCU-backed artifacts in our dataset. \texttt{per\_layer\_kernel} provides the representative LLaMA-3.1-8B/H100 prefill decomposition used for the layer-level example, while \texttt{kernels\_labeled} provides broader per-kernel NCU timing and shape records across four GPU platforms and 11 model architectures (Appendix Tables~\ref{tab:hardware} and~\ref{tab:models}).

\section{Results}
\label{sec:results}

We benchmark dense LLaMA/Qwen and MoE Mixtral/gpt-oss models spanning 8B--120B parameters across four GPU platforms. Experiments use the tensor-parallel degree required for each model to fit, and run through the closed-loop serving harness defined in Section~\ref{sec:framework}. We sweep concurrency $C$ to saturation for each workload and compare serving behavior across chat, coding, terminal-use and computer-use agentic sessions.

Section~\ref{sec:setup} summarizes the experimental setup, Section~\ref{sec:saturation_results} presents agentic workload behavior across profiles, Section~\ref{sec:rooflines} maps workloads onto the serving hardware roofline and decomposes per-component hardware bottlenecks.

\subsection{Experimental Setup}
\label{sec:setup}

We evaluate dense LLaMA/Qwen and MoE Mixtral/gpt-oss models on four GPU platforms (H100-80GB, A100-40GB, RTX~3090-24GB, and RTX~2080Ti-22GB), spanning datacenter and consumer accelerators. Each benchmark run uses only the GPU subset assigned to its tensor-parallel configuration, with no other serving workload sharing those GPUs during measurement. Serving runs use vLLM v0.19.0/v0.19.1 and SGLang v0.5.9, both driven by the same closed-loop harness through an OpenAI-compatible chat API; backend version, tensor-parallel degree, context limit, and launch configuration are recorded per run, while each backend applies the model's native tokenizer and chat template. Dense and Mixtral models use BF16; gpt-oss-20b and gpt-oss-120b use the released MXFP4 weights for MoE projections with other tensors in BF16. The serving environment uses CUDA 12.8, NVIDIA driver 13.0, PyTorch 2.10.0, and Python 3.12; Nsight Compute profiling uses ncu v2024.3.2 across A100, H100, RTX~3090, and RTX~2080Ti hosts. Appendix Tables~\ref{tab:hardware}--\ref{tab:tp_configs} provide the hardware, backend launch settings, model architectures, and tensor-parallel configurations.

\subsection{Benchmark Results}
\label{sec:saturation_results}

\begin{figure*}[t]
    \centering
    \includegraphics[width=\linewidth]{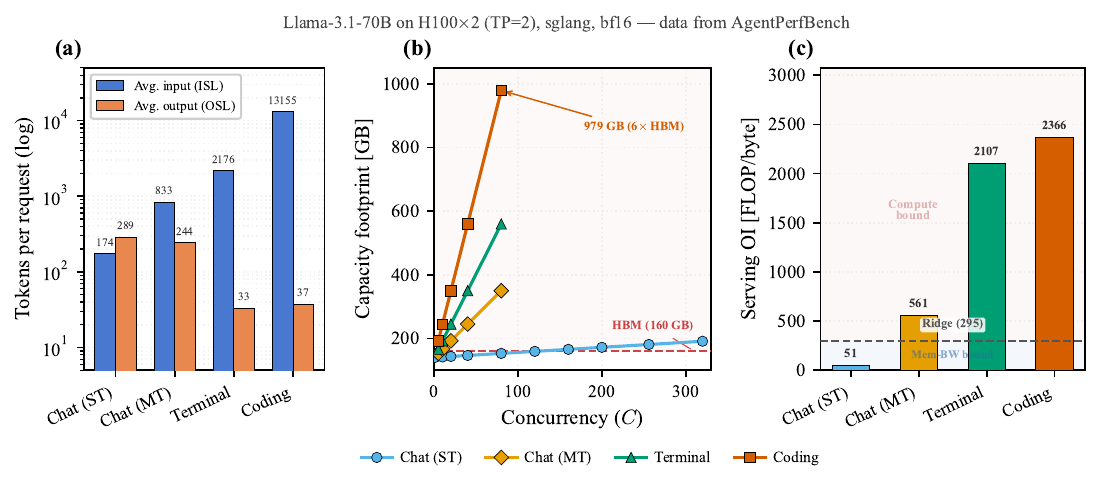}
    \caption{Agentic workload characterization (LLaMA-3.1-70B, H100$\times$2, TP=2, SGLang v0.5.9, bf16). (a)~Token usage; coding agents are prefill-dominated (ISL/OSL $\approx$\,355:1). (b)~Capacity footprint vs.\ concurrency. (c)~Serving OI; chat stays below the H100 ridge (295\,FLOP/byte), agentic workloads cross it.}
    \label{fig:agentic_contrast}
\end{figure*}

\paragraph{Workload characterization.}
Figure~\ref{fig:agentic_contrast} shows that the canonical workload profiles separate along two axes: operational intensity and capacity footprint.
Coding agents are prefill-dominated: ISL/OSL $\approx$\,355:1, compared to single-digit ratios for chat (Figure~\ref{fig:agentic_contrast}a).
The high ISL pushes coding-agent OI to 2{,}366, above the H100 ridge (295\,FLOP/byte), while chat sits at OI\,{=}\,51, a $46\times$ gap (Figure~\ref{fig:agentic_contrast}c).
On the capacity axis, long-context profiles cross the 160\,GB HBM budget of two H100s as concurrency increases, while chat remains within budget (Figure~\ref{fig:agentic_contrast}b).
Most serving benchmarks, such as InferenceX~\citep{inferencex}, collapse this space to ShareGPT-style chat or fixed-length synthetic prompts at one operating point.
This reduction misses the ISL/OSL distribution of agentic workloads even in single-turn settings; in multi-turn sessions, the gap compounds through snowballing as each turn appends prior context.
Table~\ref{tab:singleturn_latency_gap} quantifies this mismatch under a controlled single-turn comparison: matched on model, hardware, backend, tensor-parallel degree, and $C{=}80$, coding-agent planning calls have a median $2.8\times$ higher TTFT and $1.3\times$ higher TPOT than chat across 40 released \texttt{trace\_replay} groups.
Our benchmark therefore exposes not just different token counts, but different latency regimes as ISL/OSL distributions shift.

\begin{table}[t]
\caption{Single-turn coding-agent vs.\ chat serving metrics gap in released \texttt{trace\_replay} rows at $C{=}80$. Each row compares \texttt{coding-singleturn} against \texttt{chat-singleturn}; ratios are coding/chat after matching model, hardware, backend, and tensor-parallel degree, so values above $1\times$ mean the coding-agent planning call is slower.}
\label{tab:singleturn_latency_gap}
\centering
\small
\begin{adjustbox}{max width=\linewidth}
\begin{tabular}{@{}lrrr@{}}
\toprule
\textbf{Matched comparison} & \textbf{Pairs} & \textbf{TTFT (coding/chat)} & \textbf{TPOT (coding/chat)} \\
\midrule
All matched pairs (median) & 40 & $2.8\times$ & $1.3\times$ \\
Maximum TTFT ratio & 1 & $14.5\times$ & --- \\
Example: LLaMA-3.1-8B, H100, vLLM & 1 & $3.1\times$ & $2.2\times$ \\
Example: LLaMA-3.1-8B, H100$\times$2, vLLM & 1 & $3.9\times$ & $1.9\times$ \\
\bottomrule
\end{tabular}
\end{adjustbox}
\end{table}

\paragraph{Saturation behavior.}
Closed-loop concurrency sweeps show how the workload mismatch changes under load.
Table~\ref{tab:multiturn_summary} reports medium multi-turn profiles up to $C{=}80$.
On LLaMA-3.1-8B, coding agent raises median TTFT from 343\,ms for chat to 11{,}719\,ms and reduces request throughput from 13.5 to 2.4\,req/s; terminal agent raises TTFT to 6{,}889\,ms and reduces throughput to 3.0\,req/s.
The gap persists across model families: for gpt-oss-20b, chat TTFT is 267\,ms, compared with 2{,}224\,ms for coding agent and 1{,}744\,ms for terminal agent.

\begin{table*}[t]
\caption{Medium multi-turn \texttt{trace\_replay} saturation metrics on H100-SXM5-80GB with vLLM v0.19.0 and prefix caching enabled. Each cell reports the observed saturation point for one model/profile pair; TTFT and TPOT are medians in ms. Column groups use the workload names from the main text; exact profile IDs are listed in Appendix Table~\ref{tab:released_profile_ids}.}
\label{tab:multiturn_summary}
\centering
\resizebox{\textwidth}{!}{%
\begin{tabular}{@{}lc rrr rrr rrr rrr@{}}
\toprule
& & \multicolumn{3}{c}{\textbf{Chat}} & \multicolumn{3}{c}{\textbf{Coding agent}} & \multicolumn{3}{c}{\textbf{Terminal agent}} & \multicolumn{3}{c}{\textbf{Computer-use agent}} \\
\cmidrule(lr){3-5}\cmidrule(lr){6-8}\cmidrule(lr){9-11}\cmidrule(lr){12-14}
\textbf{Model} & \textbf{TP}
  & TTFT & TPOT & Req/s
  & TTFT & TPOT & Req/s
  & TTFT & TPOT & Req/s
  & TTFT & TPOT & Req/s \\
\midrule
\rowcolor{gray!15}
\multicolumn{14}{@{}l}{\textit{Dense}} \\
\midrule
LLaMA-3.1-8B   & 1 &   343 &  9.4 & 13.5 & 11719 & 227.3 & 2.4 &  6889 & 197.2 & 3.0 & 1037 & 22.8 &  6.5 \\
Qwen3.5-9B     & 1 &   953 & 12.0 & 11.0 &  6954 & 168.7 & 3.2 &  2877 &  47.1 & 4.6 & 1539 & 22.5 &  5.9 \\
Qwen3.5-27B    & 2 &  1518 & 21.5 &  6.0 &  2729 &  62.3 & 5.3 &  2519 &  52.0 & 4.7 & 2709 & 33.2 &  3.6 \\
\midrule
\rowcolor{gray!15}
\multicolumn{14}{@{}l}{\textit{MoE}} \\
\midrule
Mixtral-8$\times$7B & 2 & 269 & 18.9 &  7.1 & \multicolumn{3}{c}{---} & \multicolumn{3}{c}{---} & 1243 & 33.2 &  4.9 \\
gpt-oss-20b    & 1 &   267 &  8.8 & 15.3 &  2224 &  17.1 & 16.5 &  1744 &  15.8 & 11.6 &  699 & 14.1 & 10.1 \\
gpt-oss-120b   & 2 &   224 & 13.6 & 11.0 &  2072 &  25.5 & 13.6 &  1418 &  19.0 &  8.8 &  687 & 17.5 &  8.8 \\
\bottomrule
\end{tabular}%
}
\end{table*}

\begin{table}[t]
\centering
\caption{High-concurrency request-throughput check using matched \texttt{synthetic\_distributional} rows. Throughput columns report requests/s; $\Delta$ compares $C{=}320$ to $C{=}200$.}
\label{tab:synthetic_high_c}
\small
\begin{adjustbox}{max width=\linewidth}
\begin{tabular}{@{}llrrr@{}}
\toprule
\textbf{Profile} & \textbf{Config} & \textbf{Req/s @ $C{=}200$} & \textbf{Req/s @ $C{=}320$} & \textbf{$\Delta$} \\
\midrule
Chat single-turn & LLaMA-3.1-8B, H100, vLLM & 9.6 & 31.2 & $+224\%$ \\
Chat single-turn & LLaMA-3.1-8B, A100$\times$4, SGLang & 11.2 & 17.0 & $+52\%$ \\
Chat multi-turn & Mixtral-8$\times$7B, H100$\times$4, vLLM & 27.5 & 36.5 & $+32\%$ \\
\midrule
Coding agent & gpt-oss-20b, RTX~3090$\times$2, vLLM & 9.2 & 2.2 & $-76\%$ \\
Terminal agent & Qwen3.5-9B, A100$\times$4, vLLM & 21.8 & 9.5 & $-57\%$ \\
Terminal agent & Mixtral-8$\times$7B, H100$\times$4, vLLM & 30.9 & 17.6 & $-43\%$ \\
Computer-use agent & LLaMA-3.1-8B, A100$\times$4, vLLM & 16.7 & 8.3 & $-51\%$ \\
\bottomrule
\end{tabular}
\end{adjustbox}
\end{table}

The released \texttt{trace\_replay} split stops at $C{=}80$, so we use \texttt{synthetic\_distributional} rows to test higher concurrency. Matching by profile, model, hardware, backend, and tensor-parallel degree, Table~\ref{tab:synthetic_high_c} compares request throughput at $C{=}200$ and $C{=}320$. Across 63 comparable chat groups, median request throughput grows $+11\%$, with 33 groups still improving by more than $10\%$. In contrast, the 66 comparable agentic groups are mostly already at or past saturation by $C{=}200$: 41 remain within $\pm10\%$ at $C{=}320$, while 15 lose more than $10\%$ throughput. Chat workloads often continue benefiting from larger batches, while agentic workloads plateau or degrade under KV-capacity pressure. Single-operating-point benchmarks that do not cover agentic serving regimes or sweep beyond saturation cannot expose this divergence between chat and agentic scaling behavior.

\subsection{Roofline Bottleneck Diagnosis}
\label{sec:rooflines}

Different agentic workloads occupy distinct roofline regimes:
coding-agent OI reaches $2{,}409$ while chat peaks at $181$ on
the same H100, a $13\times$ separation
(Appendix Figure~\ref{fig:roofline_multiturn}).
OI and CF depend only on model architecture and workload token
statistics, so regime classification requires no serving run and extends to hardware not yet deployed:
\begin{itemize}[nosep,leftmargin=*]
  \item the ridge point (peak FLOP/s divided by peak bandwidth;
        295\,FLOP/byte for H100) classifies each workload as
        compute-bound or bandwidth-bound, identifying which
        hardware resource limits throughput
        (Table~\ref{tab:oi_eff});
  \item HBM capacity determines at what concurrency KV-cache
        growth saturates memory, bounding the maximum batch size
        before any request is served
        (Appendix Figure~\ref{fig:roofline_multiturn}b).
\end{itemize}

Appendix Figure~\ref{fig:roofline_multiturn} plots this separation for LLaMA-3.1-8B on a single H100, empirically validating the regime classification predicted by \citet{zhao2026heterogeneous}.

\paragraph{OI profile separation.} The H100 ridge point, $\text{OI}_{\text{ridge}} = 989\,\text{TFLOP/s}\;/\;3{,}352\,\text{GB/s} = 295$\,FLOP/byte, separates memory-bandwidth-bound from compute-bound execution.
Chat stays below it, peaking at $\text{OI}_{\text{eff}} = 181$ (Table~\ref{tab:oi_eff}), remaining memory-bandwidth-bound across the full concurrency sweep. 
The other agentic workloads cross the ridge: computer-use reaches $\text{OI}_{\text{eff}} = 886$ ($3.0\times$ the ridge), terminal reaches $1{,}809$ ($6.1\times$), and coding agent reaches $2{,}409$ ($8.2\times$) at $C = 80$, all compute-bound.
Peak separation between coding agent and chat is $13\times$. The mechanism is the snowballing effect (Section~\ref{sec:background}): context accumulation across turns grows ISL, shifting prefill toward large General Matrix Multiplications (GEMMs) whose arithmetic intensity exceeds the ridge.

\paragraph{Capacity footprint saturation.} In the analytical CF model used for Appendix Figure~\ref{fig:roofline_multiturn}b, LLaMA-3.1-8B BF16 weights contribute 16\,GB on a single H100, leaving a nominal 64\,GB KV-cache budget before the 80\,GB HBM wall.
At $C=80$, chat remains compact at 26\,GB, and computer use reaches 65\,GB. The longer agentic profiles exceed the single-H100 HBM budget: coding agent exceeds the wall by $C=40$ and reaches $155\,GB$ at $C=80$, while terminal reaches $126\,GB$ (Appendix Figure~\ref{fig:roofline_multiturn}b clips both at the 80\,GB wall).
Long agentic workloads therefore face dual pressure: 1) compute-bound execution on the OI axis and 2) memory-capacity pressure on the CF axis, where the growing KV cache saturates HBM.
Table~\ref{tab:synthetic_high_c} supports this predicted CF pressure: from $C{=}200$ to $C{=}320$, representative agentic profiles lose 43--76\% throughput while chat profiles continue scaling.
Aggregate OI does not isolate which layer components create these bottlenecks.

\begin{figure}[!ht]
    \centering
    \includegraphics[width=0.75\linewidth]{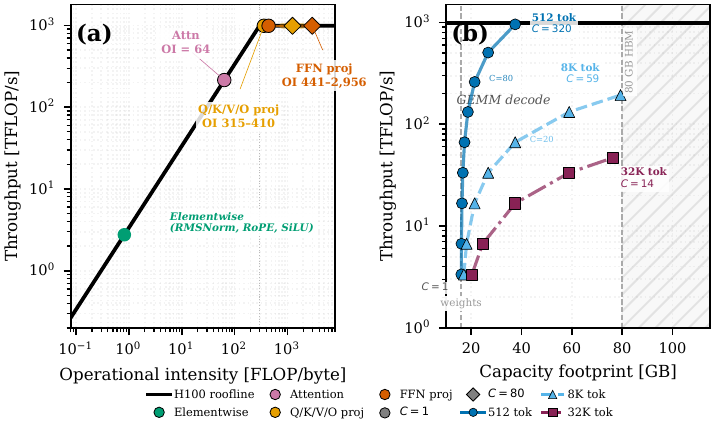}
    \caption{Per-component roofline decomposition (LLaMA-3.1-8B, H100, BF16, TP=1). (a)~Analytical prefill OI; circles mark $C{=}1$, diamonds $C{=}80$. Projection GEMMs sit above the ridge (295\,FLOP/byte); elementwise operations and attention remain bandwidth-bound at both concurrencies. (b)~Decode-side throughput ceiling versus capacity footprint for 512-token, 8K-token, and 32K-token contexts. Short contexts remain within HBM at high concurrency, while 32K-token contexts hit the 80\,GB capacity wall at $C{=}14$.}
    \label{fig:per-layer-oi-cf}
\end{figure}

\paragraph{Per-component bottleneck diagnosis.}
\label{sec:kernel}
A single serving OI hides the 8--28$\times$ gap between prefill
and decode (Table~\ref{tab:phase_oi}), and a single CF number
does not show when KV capacity overrides the component ceilings.
Appendix Table~\ref{tab:per-component-oi} and Figure~\ref{fig:per-layer-oi-cf}
decompose one LLaMA-3.1-8B decoder layer into per-component OI,
modeled latency share, and the limiting regime for both phases.
Because this diagnostic uses model architecture plus the target
accelerator's peak compute, bandwidth, and HBM capacity, the same
calculation can be re-run for hardware that has not yet been
provisioned.

\paragraph{Projection and elementwise OI.}
At $C{=}80$ with 512-token inputs, prefill pushes $512 \times 80 = 40{,}960$ tokens through each layer versus 80 for decode (Section~\ref{sec:background}), placing every prefill projection above the H100 ridge of 295\,FLOP/byte while every decode projection remains below it (Appendix Table~\ref{tab:per-component-oi}).
Attention and elementwise operations are bandwidth-bound in both phases (OI\,$\leq$\,64).
The prefill projection aggregate is therefore compute-bound and accounts for 81.8\% of the modeled layer latency; the decode projection aggregate (OI\,{=}\,78 at $C{=}80$) remains bandwidth-bound because no component crosses the ridge.
As context accumulates across turns, prefill ISL grows and pushes projection OI further above the ridge.
Decode projection OI is set by concurrency alone, regardless of context length.
At $C{=}80$ the resulting prefill-to-decode OI gap (the kernel-level source of the regime separation identified above) ranges from $8\times$ (chat) to $28\times$ (coding agent) across workloads (Appendix Table~\ref{tab:phase_oi}).

\paragraph{KV-capacity pressure.}
At short context (512 tokens), KV per sequence is small enough that high concurrency fits within the 80\,GB HBM budget. At long context (32K tokens), KV per sequence dominates HBM and the capacity wall is reached at $C{\approx}14$ (Appendix Table~\ref{tab:per-component-oi}). Once CF exceeds HBM, capacity is the phase-level bottleneck: the per-component Compute/BW labels describe the in-memory limiting resource, but the actual serving step must reduce batch size, page/evict KV blocks, or recompute cache state, so the accelerator is generally underutilized rather than cleanly compute- or bandwidth-saturated. Together with prefill projection GEMMs compute-bound at high concurrency, this capacity override completes the per-component explanation of the dual pressure observed in the aggregate roofline.

\subsection{Discussion}
\label{sec:discussion}

Three limitations bound the current evaluation. First, all results are produced on single-node multi-GPU servers with tensor parallelism up to TP=8.
Multi-node serving can introduce cross-node communication and, under pipeline parallelism, inter-stage bubbles that are not captured by tensor-parallel single-node measurements.
A pipeline-parallel concurrency sweep will quantify this overhead.
Second, multi-turn sessions are bounded by the per-run context cap (e.g., 32K tokens for LLaMA-3.1-8B long-context runs).
Sessions that exceed this cap are truncated, so the snowballing effect reported here is a lower bound on what 128K-context deployments will exhibit.
Third, the current benchmark records logical prefix-cache estimates but does not directly measure server-side KV eviction, context compaction, or cold-cache recomputation.
Adding those mechanisms to long multi-turn runs may expose additional latency beyond the measurements reported here.

As Section~\ref{sec:rooflines} shows, the $13\times$ peak OI
separation between coding-agent and chat workloads places them on
opposite sides of the roofline ridge; the wider this gap across a
deployment's workload mix, the stronger the case for
prefill-decode disaggregation.

The 8--28$\times$ prefill-to-decode OI gap across the workload surface (Table~\ref{tab:phase_oi}), combined with CF saturation at $C \geq 80$, makes these workloads natural candidates for prefill-decode disaggregation (PDD)~\citep{distserve,splitwise}: unified serving forces compute-bound prefill and memory-bandwidth-bound decode onto the same hardware, and the per-phase gap is wide enough that separation should yield measurable throughput gains.

\section{Conclusion}
\label{sec:conclusion}

Agentic LLM workloads occupy a hardware regime that no prior inference benchmark captures.
Across 22 workload profiles, 9 models, and 14 GPU configurations (3{,}197 sweep rows, 148{,}077 NCU records), \textsc{AgentPerfBench} identifies a dual pressure that separates agentic serving from chat.

On a single H100, coding-agent OI reaches $2{,}409$ versus $181$ for chat, a $13\times$ separation that places agentic prefill above the ridge point while chat remains bandwidth-bound (Section~\ref{sec:rooflines}, Figure~\ref{fig:roofline_multiturn}).
The per-component decomposition reveals why: projection GEMMs become compute-bound in prefill (OI up to $2{,}956$ at $C{=}80$), while attention, elementwise, and all decode components stay bandwidth-bound (Table~\ref{tab:per-component-oi}).
The resulting prefill-to-decode OI gap spans 8--28$\times$ across workloads (Table~\ref{tab:phase_oi}).
Simultaneously, KV-cache growth creates capacity pressure: coding-agent sessions exceed the 80\,GB HBM wall by $C{=}40$, and saturation sweeps show agentic profiles lose 43--76\% throughput from $C{=}200$ to $C{=}320$ while chat continues scaling (Table~\ref{tab:synthetic_high_c}).

The dual pressure, compute-bound prefill and capacity-constrained decode on opposite sides of the roofline ridge, makes agentic workloads natural candidates for prefill--decode disaggregation (Section~\ref{sec:discussion}); because the roofline analysis requires only model architecture and published hardware specs, it extends to accelerators not yet deployed.
Three directions remain: multi-node pipeline-parallel serving where cross-node communication may shift the bottleneck balance, explicit measurement of KV eviction, compaction, and recomputation under long multi-turn sessions, and extension to 128K+ context windows where the snowballing effect reported here is a lower bound.

\section{Dataset and Reproducibility}
\label{sec:dataset}

We release five Hugging Face dataset configurations:
\begin{itemize}
    \item \textbf{Serving summaries}: 3{,}197 main sweep rows across 9 served models, 14 GPU configurations, and two engines (vLLM v0.19.0 and SGLang v0.5.9). \texttt{trace\_replay} contains 2{,}932 rows that replay source ISL/OSL sequences across 17 profiles; \texttt{synthetic\_distributional} contains 265 rows sampled from fitted real-workload statistics across 5 profiles.
    \item \textbf{Layer validation}: \texttt{per\_layer\_kernel} contains 37 rows for the LLaMA-3.1-8B/H100 prefill-phase decomposition used in Section~\ref{sec:kernel}, including analytical component rows and NCU-backed kernel measurements.
    \item \textbf{Kernel profiles}: \texttt{kernels\_labeled} contains 148{,}077 per-kernel Nsight Compute (NCU) records across A100, H100, RTX~3090, and RTX~2080Ti platforms and 13 model/sweep sources. Rows include timing, matrix-shape, operation-family, memory-traffic, and launch-configuration fields for downstream kernel-latency modeling.
    \item \textbf{Replay validation}: \texttt{mse\_validation} contains 28 H100/LLaMA-3.1-8B/vLLM rows used to validate the distributional synthetic replay generator, including paired synthetic and real \texttt{trace\_replay} runs plus retained debug rows.
    \item \textbf{Benchmark tool}: Open-source benchmarking and profiling scripts for closed-loop concurrency sweeps against OpenAI-compatible endpoints, with warmup, success filtering, TTFT, TPOT, ITL, E2EL, throughput, and per-run metadata collection.
\end{itemize}

All artifacts are released under an open-source license with a Croissant metadata record for machine-readable discovery.\footnote{Hosted at \url{[anonymized for review]}.}

\bibliographystyle{plainnat}
\bibliography{references}

\appendix

\section{Experimental Details}
\label{app:experimental_details}

\begin{table}[h]
\caption{GPU platforms used in released runs.}
\label{tab:hardware}
\centering
\footnotesize
\setlength{\tabcolsep}{2pt}
\renewcommand{\arraystretch}{1.12}
\begin{adjustbox}{max width=\linewidth}
\begin{tabular}{@{}lcccc@{}}
\toprule
 & \makecell{\textbf{A100-SXM4}\\\textbf{40GB}} & \makecell{\textbf{H100-SXM5}\\\textbf{80GB}} & \makecell{\textbf{RTX 3090}\\\textbf{24GB}} & \makecell{\textbf{RTX 2080Ti}\\\textbf{22GB}} \\
\midrule
Available GPUs & 8$\times$ & \makecell{8$\times$ on-premise\\RunPod nodes} & 8$\times$ & 8$\times$ \\
\addlinespace[1pt]
Benchmarked TP & 1, 2, 4, 8 & 1, 2, 4 & 1, 2, 4, 8 & 1, 2, 4 \\
\addlinespace[1pt]
Interconnect  & NVLink/NVSwitch & NVLink/NVSwitch & PCIe & PCIe \\
\addlinespace[1pt]
Run source & on-premise & on-premise + RunPod & on-premise & on-premise \\
\addlinespace[1pt]
Tensor peak & \makecell{312 TFLOPS\\BF16} & \makecell{989 TFLOPS\\BF16} & \makecell{71 TFLOPS\\FP16} & \makecell{108 TFLOPS\\FP16} \\
\addlinespace[1pt]
Memory bandwidth & \makecell{1{,}555 GB/s\\HBM2} & \makecell{3{,}352 GB/s\\HBM3} & \makecell{936 GB/s\\GDDR6X} & \makecell{616 GB/s\\GDDR6} \\
\bottomrule
\end{tabular}
\end{adjustbox}
\end{table}

\paragraph{Serving flags.}
The launch scripts start one backend server per sweep row, run all requested profiles and concurrencies against that server, then tear it down. The sweep manifest records backend, tensor-parallel degree, context limit, and memory fraction for each cell. Table~\ref{tab:serving_flags} lists the launch settings used by both serving engines; vLLM-only scheduling and CUDA-graph flags are not treated as SGLang settings. SGLang model-specific extras observed in the manifests include \texttt{--trust-remote-code}, \texttt{--disable-overlap-schedule}, and \texttt{--disable-piecewise-cuda-graph}.

\begin{table}[h]
\caption{Backend launch-setting summary from the released serving summaries and sweep manifests. We report knobs that affect scheduling, memory, context length, or model loading; fixed host, port, API-key, and model-path flags are omitted.}
\label{tab:serving_flags}
\centering
\scriptsize
\setlength{\tabcolsep}{3pt}
\renewcommand{\arraystretch}{1.08}
\begin{adjustbox}{max width=\linewidth}
\begin{tabular}{@{}p{0.15\linewidth}p{0.25\linewidth}p{0.25\linewidth}p{0.27\linewidth}@{}}
\toprule
\textbf{Setting} & \textbf{vLLM} & \textbf{SGLang} & \textbf{Scope / notes} \\
\midrule
Entrypoint & OpenAI API server & SGLang launch server & One server per sweep row \\
Tensor parallelism & \texttt{--tensor-parallel-size} & \texttt{--tp} & Manifest TP value \\
Precision & \texttt{--dtype bfloat16} / \texttt{auto} & \texttt{--dtype bfloat16} / \texttt{auto} & \texttt{auto} appears for gpt-oss-120b launch attempts \\
Memory budget & \texttt{--gpu-memory-utilization} & \texttt{--mem-fraction-static} & Manifest values span roughly 0.80--0.95 \\
Context limit & \texttt{--max-model-len} & \texttt{--context-length} & Manifest values span 4K--32K tokens \\
Prefix reuse & \texttt{--enable-prefix-caching} & no explicit launch flag in manifests & Client metadata records prefix-cache estimates \\
Chunked prefill & \texttt{--enable-chunked-prefill} & no explicit launch flag in manifests & SGLang rows record this as unknown in client metadata \\
Memory-safety flags & \makecell[l]{\texttt{--enforce-eager}\\\texttt{--disable-custom-all-reduce}} & --- & A100/tight-memory vLLM only \\
Model-specific extras & \texttt{--trust-remote-code} & \makecell[l]{\texttt{--trust-remote-code}\\backend-specific disable flags} & Model/backend specific \\
\bottomrule
\end{tabular}
\end{adjustbox}
\end{table}

\begin{table}[h]
\caption{Model configurations evaluated. Dense models use grouped-query attention (GQA); MoE models use top-$k$ expert routing. Models marked with a dagger ($^\dagger$) are included only in the NCU kernel profiling dataset (\texttt{kernels\_labeled}) and were not run through the serving benchmark.}
\label{tab:models}
\centering
\small
\begin{tabular}{lcrrrrrrcc}
\toprule
\textbf{Model} & \textbf{Type} & \textbf{Params} & \textbf{$d_\text{model}$} & \textbf{Layers} & \textbf{Heads} & \textbf{KV} & \textbf{$d_\text{ffn}$} & \textbf{Experts} & \textbf{top-$k$} \\
\midrule
LLaMA-3.1-8B        & Dense & 8B   & 4{,}096 & 32 & 32 & 8 & 14{,}336 & ---  & --- \\
LLaMA-3.1-70B       & Dense & 70B  & 8{,}192 & 80 & 64 & 8 & 28{,}672 & ---  & --- \\
LLaMA-3.3-70B       & Dense & 70B  & 8{,}192 & 80 & 64 & 8 & 28{,}672 & ---  & --- \\
Qwen2.5-72B         & Dense & 72B  & 8{,}192 & 80 & 64 & 8 & 29{,}568 & ---  & --- \\
Qwen3.5-9B          & Dense & 9B   & 4{,}096 & 32 & 16 & 4 & 12{,}288 & ---  & --- \\
Qwen3.5-27B         & Dense & 27B  & 5{,}120 & 64 & 24 & 4 & 17{,}408 & ---  & --- \\
Mixtral-8$\times$7B & MoE   & 47B  & 4{,}096 & 32 & 32 & 8 & 14{,}336 & 8    & 2 \\
gpt-oss-20b         & MoE   & 20B  & 2{,}880 & 24 & 64 & 8 & 2{,}880 & 32   & 4 \\
gpt-oss-120b        & MoE   & 120B & 2{,}880 & 36 & 64 & 8 & 2{,}880 & 128  & 4 \\
\midrule
\multicolumn{10}{l}{\footnotesize\textit{NCU profiling only (not included in serving benchmarks):}} \\
Gemma-2-9B$^\dagger$  & Dense & 9B   & 3{,}584 & 42 & 16 & 8 & 14{,}336 & ---  & --- \\
Granite-3.0-8B$^\dagger$ & Dense & 8B & 4{,}096 & 40 & 32 & 8 & 12{,}800 & ---  & --- \\
\bottomrule
\end{tabular}
\end{table}

\begin{table}[h]
\caption{Tensor-parallel configurations actually run in the released serving summaries. Entries list observed TP degrees after collapsing multi-GPU labels within each platform family. gpt-oss models use MXFP4 weights; all others bfloat16.}
\label{tab:tp_configs}
\centering
\small
\begin{tabular}{lcccc}
\toprule
\textbf{Model} & \textbf{2080Ti} & \textbf{RTX 3090} & \textbf{A100-40GB} & \textbf{H100} \\
\midrule
LLaMA-3.1-8B        & 1, 2, 4 & 1, 2, 4, 8 & 1, 2, 4, 8 & 1, 2, 4 \\
LLaMA-3.1-70B       & ---     & 8          & 4, 8       & 1, 2, 4 \\
LLaMA-3.3-70B       & ---     & ---        & 4, 8       & 1, 2, 4 \\
Qwen2.5-72B         & ---     & ---        & 4, 8       & 1, 2, 4 \\
Mixtral-8$\times$7B & ---     & 8          & 4, 8       & 2, 4 \\
Qwen3.5-9B          & 2, 4    & 1, 2, 4, 8 & 1, 2, 4, 8 & 1, 2 \\
Qwen3.5-27B         & ---     & 4          & 8          & 2 \\
gpt-oss-20b         & ---     & 1, 2, 4    & 1, 2, 4, 8 & 1, 2, 4 \\
gpt-oss-120b        & ---     & ---        & 8          & 2 \\
\bottomrule
\end{tabular}
\end{table}

\begin{table}[h]
\caption{Released workload profile identifiers corresponding to the workload names in Table~\ref{tab:profiles}. Rows are grouped by released dataset split: \texttt{synthetic\_distributional} contains synthetic multi-turn profiles and canonical single-turn baselines, while \texttt{trace\_replay} contains archived replay profiles from recorded agent sessions and ShareGPT conversations, plus synthetic stress profiles. Source basis identifies the conversation, trajectory, or empirical-statistic source used to construct each profile. Turns reports 1 for single-turn rows, medians for synthetic multi-turn rows, and turn-count ranges for \texttt{trace\_replay} multi-turn profiles. ISL and OSL report profile-level source medians.}
\label{tab:released_profile_ids}
\centering
\scriptsize
\begin{adjustbox}{max width=\linewidth}
\begin{tabular}{@{}lllllrr@{}}
\toprule
\textbf{Workload} & \textbf{Profile ID} & \textbf{Source basis} & \textbf{Session} & \textbf{Turns} & \textbf{ISL} & \textbf{OSL} \\
\midrule
\rowcolor{gray!15}
\multicolumn{7}{@{}l}{\textit{Synthetic distributional}} \\
\midrule
Single-turn chat & \texttt{chat-singleturn} & ShareGPT conversations & single-turn & 1 & 187 & 299 \\
Coding planning call & \texttt{coding-singleturn} & SWE-Bench prompts & single-turn & 1 & 6{,}300 & 280 \\
Chatbot session & \texttt{chat-multiturn} & ShareGPT summaries & multi-turn & 10 & 1{,}291 & 169 \\
Coding-agent session & \texttt{swebench-multiturn} & SWE-Bench statistics & multi-turn & 85 & 9{,}995 & 32 \\
Terminal-agent session & \texttt{terminalbench-multiturn} & TerminalBench statistics & multi-turn & 61 & 7{,}811 & 31 \\
Computer-use session & \texttt{osworld-multiturn} & OSWorld statistics & multi-turn & 8 & 1{,}399 & 85 \\
\midrule
\rowcolor{gray!15}
\multicolumn{7}{@{}l}{\textit{Trace replay}} \\
\midrule
Single-turn chat & \texttt{chat-singleturn} & ShareGPT conversations & single-turn & 1 & 187 & 299 \\
Single-turn chat & \texttt{chat-short} & ShareGPT conversations & single-turn & 1 & 122 & 168 \\
Single-turn chat & \texttt{chat-medium} & ShareGPT conversations & single-turn & 1 & 141 & 288 \\
Coding planning call & \texttt{coding-singleturn} & SWE-Bench prompts & single-turn & 1 & 6{,}300 & 280 \\
Prefill stress & \texttt{prefill-heavy} & Random tokens & single-turn & 1 & 9{,}223 & 250 \\
Decode stress & \texttt{decode-heavy} & Random tokens & single-turn & 1 & 300 & 353 \\
Random-token stress & \texttt{random-1k} & Random tokens & single-turn & 1 & 1{,}160 & 301 \\
Chatbot session & \texttt{chat-multiturn-short} & ShareGPT conversations & multi-turn & 3--5 & 679 & 298 \\
Chatbot session & \texttt{chat-multiturn-medium} & ShareGPT conversations & multi-turn & 5--10 & 839 & 246 \\
Chatbot session & \texttt{chat-multiturn-long} & ShareGPT conversations & multi-turn & 10--20 & 938 & 150 \\
Coding-agent session & \texttt{swebench-multiturn-short} & SWE-Bench trajectories & multi-turn & 13--30 & 3{,}530 & 28 \\
Coding-agent session & \texttt{swebench-multiturn-medium} & SWE-Bench trajectories & multi-turn & 30--80 & 7{,}299 & 32 \\
Terminal-agent session & \texttt{terminalbench-multiturn-short} & TerminalBench trajectories & multi-turn & 2--20 & 1{,}994 & 29 \\
Terminal-agent session & \texttt{terminalbench-multiturn-medium} & TerminalBench trajectories & multi-turn & 20--60 & 4{,}956 & 33 \\
Computer-use session & \texttt{osworld-multiturn-short} & OSWorld trajectories & multi-turn & 2--10 & 1{,}343 & 82 \\
Computer-use session & \texttt{osworld-multiturn-medium} & OSWorld trajectories & multi-turn & 10--20 & 1{,}366 & 85 \\
Computer-use session & \texttt{osworld-multiturn-long} & OSWorld trajectories & multi-turn & 20--30 & 1{,}388 & 85 \\
\bottomrule
\end{tabular}
\end{adjustbox}
\end{table}

\begin{figure}[h]
    \centering
    \includegraphics[width=0.82\linewidth]{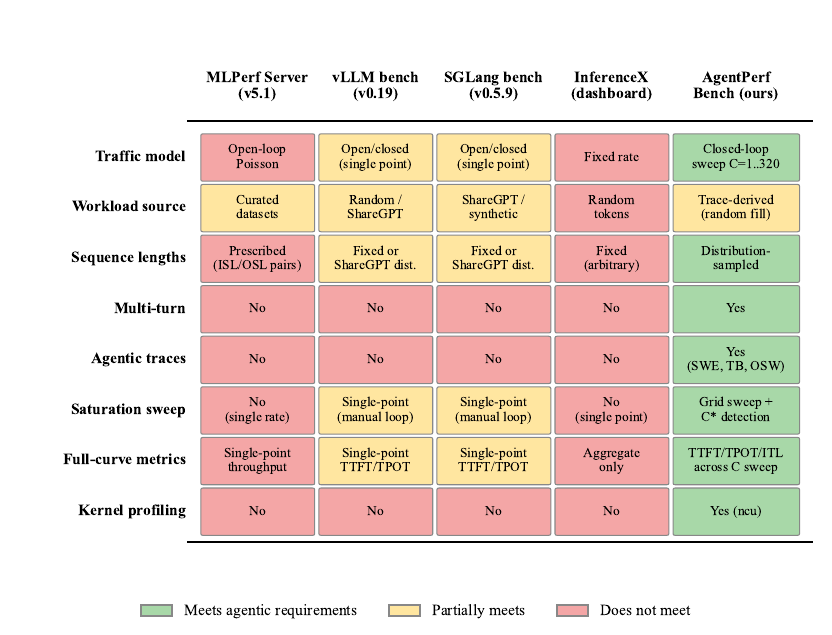}
    \caption{Feature coverage of five benchmark harnesses across agentic measurement dimensions.}
    \label{fig:methodology_comparison}
\end{figure}

\paragraph{Benchmark harness coverage.}
Figure~\ref{fig:methodology_comparison} summarizes the agentic measurement dimensions covered by representative benchmark harnesses.

\section{Additional Results}
\label{app:results}

\subsection{Distributional Replay Validation Ablation}
\label{sec:appendix_dist_replay_validation}
Table~\ref{tab:dist_replay_validation} shows the source-locked SWE-Bench ablation used to choose the distributional replay configuration. A synthetic filler with code-like token structure still has $31.1\%$ deep-turn E2EL MAPE when each session carries an independent prefix. Disabling APC reduces this to $11.7\%$, showing that prefix caching amplifies the mismatch. Adding a shared $1024$-token synthetic prefix brings deep-turn E2EL MAPE to $3.2\%$ and aggregate E2EL MAPE to $2.4\%$, within the observed noise scale. Hence synthetic distributional profiles in this paper use this prefix-aware configuration.

\begin{table}[h]
\caption{\textbf{Distributional replay validation ablation.} Source-locked SWE-Bench C=5 validation on H100/LLaMA-3.1-8B/vLLM. Errors report MAPE between synthetic and real trace replay; the turn 10--19 bin isolates the deep-context regime where prompt snowballing and APC pressure are strongest.}
\label{tab:dist_replay_validation}
\centering
\scriptsize
\begin{adjustbox}{max width=\linewidth}
\begin{tabular}{@{}llllrr@{}}
\toprule
\textbf{Replay variant} & \textbf{Tokenizer} & \textbf{Text shape} & \textbf{APC prefix} &
\textbf{Agg. E2EL MAPE} & \textbf{Turn 10--19 E2EL MAPE} \\
\midrule
Code-like baseline & real & code-like & independent & $21.6\%$ & $31.1\%$ \\
APC-off ablation & real & code-like & disabled & $9.3\%$ & $11.7\%$ \\
Prefix-aware replay & real & code-like & shared 1024-token & $2.4\%$ & $3.2\%$ \\
\bottomrule
\end{tabular}
\end{adjustbox}
\end{table}

\subsection{Released Splits and Roofline Values}
\label{sec:appendix_sweeps}
The released dataset contains two serving-summary splits: \texttt{trace\_replay}, which replays source traces and stress profiles, and \texttt{synthetic\_distributional}, which samples from fitted or empirical workload statistics. Table~\ref{tab:released_profile_ids} lists the released profile identifiers for both splits.

\begin{figure}[t]
    \centering
    \includegraphics[width=\columnwidth]{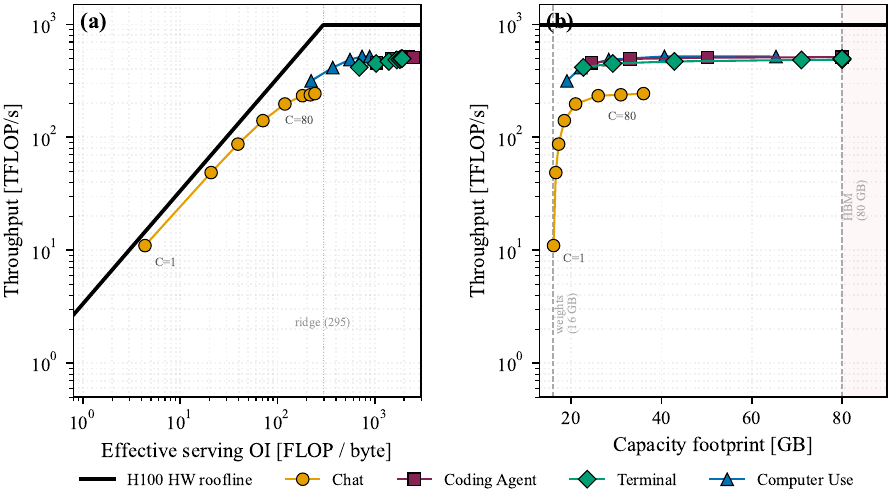}
    \caption{Multi-turn serving roofline (LLaMA-3.1-8B, H100, TP=1, vLLM v0.19, bf16, $C{=}1$--$80$). (a)~Serving OI vs.\ throughput; agentic workloads cross the H100 ridge (295\,FLOP/byte). (b)~Capacity footprint vs.\ the 80\,GB HBM wall.}
    \label{fig:roofline_multiturn}
\end{figure}

\begin{table}[h]
\caption{Effective serving OI at $C{=}80$ for multi-turn profiles (LLaMA-3.1-8B, H100, TP=1, vLLM, bf16). Values correspond to Figure~\ref{fig:roofline_multiturn}a.}
\label{tab:oi_eff}
\centering
\small
\begin{tabular}{@{}lrr@{}}
\toprule
\textbf{Profile} & $\textbf{OI}_{\textbf{eff}}$ & \textbf{Ridge multiple} \\
\midrule
Chat multi-turn & 181 & $0.6\times$ \\
Computer-use multi-turn & 886 & $3.0\times$ \\
Terminal multi-turn & 1{,}809 & $6.1\times$ \\
Coding-agent multi-turn & 2{,}409 & $8.2\times$ \\
\bottomrule
\end{tabular}
\end{table}

\begin{table}[h]
\caption{Per-phase OI gap at $C{=}80$ (LLaMA-3.1-8B, H100, BF16, TP=1). $\text{OI}_{\text{prefill}}$ is the per-layer GEMM aggregate at $L{=}\text{ISL}$ tokens; $\text{OI}_{\text{decode}}$ is the GEMM aggregate at decode batch $L{=}C{=}80$ tokens. The H100 ridge is 295\,FLOP/byte; prefill is compute-bound for all profiles, decode is bandwidth-bound.}
\label{tab:phase_oi}
\centering
\small
\begin{tabular}{@{}lrrrr@{}}
\toprule
\textbf{Profile} & \textbf{ISL} & $\textbf{OI}_{\textbf{prefill}}$ & $\textbf{OI}_{\textbf{decode}}$ & \textbf{Gap} \\
\midrule
Chat multi-turn & 833 & 634 & 78 & $8\times$ \\
Computer-use multi-turn & 4{,}666 & 1{,}695 & 78 & $22\times$ \\
Terminal multi-turn & 10{,}472 & 2{,}123 & 78 & $27\times$ \\
Coding-agent multi-turn & 12{,}929 & 2{,}208 & 78 & $28\times$ \\
\bottomrule
\end{tabular}
\end{table}

\begin{table}[h]
\centering
\caption{Layer-level bottleneck diagnosis for LLaMA-3.1-8B on H100 (BF16, TP=1). Panel (a) reports per-component OI at $C{=}80$ for a 512-token context and analytical prefill roofline-latency share; BW denotes memory-bandwidth-bound. Panel (b) is the CF override: if KV cache exceeds the 80\,GB HBM budget, capacity becomes the phase-level bound and the Compute/BW labels in panel (a) apply only to the ideal in-memory case.}
\label{tab:per-component-oi}
\scriptsize
\setlength{\tabcolsep}{2pt}
\renewcommand{\arraystretch}{0.95}
\begin{minipage}[t]{0.67\linewidth}
\centering
\textbf{(a) Component ceilings}\\[-1pt]
\begin{adjustbox}{max width=\linewidth}
\begin{tabular}{@{}lrrccr@{}}
\toprule
Component & Prefill OI & Decode OI & Prefill & Decode & Share \\
\midrule
Norms (2$\times$) & 1 & 1 & BW & BW & 1.8\% \\
Q projection & 1{,}950 & 77 & Compute & BW & 6.3\% \\
K/V projections & 803 & 73 & Compute & BW & 3.2\% \\
RoPE & 1 & 1 & BW & BW & 1.4\% \\
Attention & 64 & $\approx$0 & BW & BW & 7.2\% \\
O projection & 1{,}950 & 77 & Compute & BW & 6.3\% \\
Gate/Up projections & 2{,}956 & 78 & Compute & BW & 44.0\% \\
SiLU + Gate$\times$Up & $\leq$1 & $\leq$1 & BW & BW & 7.9\% \\
Down projection & 2{,}956 & 78 & Compute & BW & 22.0\% \\
\midrule
\textbf{Projection agg.} & \textbf{2{,}500} & \textbf{78} & \textbf{Compute} & \textbf{BW} & \textbf{81.8\%} \\
\bottomrule
\end{tabular}
\end{adjustbox}
\end{minipage}\hfill
\begin{minipage}[t]{0.30\linewidth}
\centering
\textbf{(b) Capacity override}\\[-1pt]
\begin{adjustbox}{max width=\linewidth}
\begin{tabular}{@{}lccc@{}}
\toprule
Context & KV/seq & HBM limit & $C{=}80$ \\
\midrule
512 & 0.075\,GB & $C{\approx}850$ & fits \\
8K & 1.14\,GB & $C{\approx}56$ & CF \\
32K & 4.43\,GB & $C{\approx}14$ & CF \\
128K & 17.7\,GB & $C{\approx}3$ & CF \\
\bottomrule
\end{tabular}
\end{adjustbox}
\end{minipage}

\end{table}

\end{document}